\documentclass{article}

\usepackage{arxiv}

\usepackage[utf8]{inputenc} 
\usepackage[T1]{fontenc}    
\usepackage{hyperref}       
\usepackage{url}            
\usepackage{booktabs}       
\usepackage{amsfonts}       
\usepackage{nicefrac}       
\usepackage{microtype}      
\usepackage{lipsum}
\usepackage{graphicx}
\usepackage{caption}
\graphicspath{{./}}

\title{Attention is all you need for Videos: Self-attention based Video Summarization using Universal Transformers}

\author{
  Manjot Bilkhu\\
  \texttt{mbilkhu@ucsd.edu} \\
  \texttt{UC San Diego} \\
   \And
   Siyang Wang\\
  \texttt{siw030@ucsd.edu} \\
  \texttt{UC San Diego} \\
   \And
   Tushar Dobhal\\
  \texttt{tdobhal@ucsd.edu} \\
  \texttt{UC San Diego}  
}

\begin{document}
\maketitle{}

\begin{abstract}
Video Captioning and Summarization have become very popular in the recent years due to advancements in Sequence Modelling, with the resurgence of Long-Short Term Memory networks (LSTMs) and introduction of Gated Recurrent Units (GRUs). Existing architectures extract spatio-temporal features using CNNs and utilize either GRUs or LSTMs to model dependencies with soft attention layers. These attention layers do help in attending to the most prominent features and improve upon the recurrent units, however, these models suffer from the inherent drawbacks of the recurrent units themselves. The introduction of the Transformer model has driven the Sequence Modelling field into a new direction. In this project, we implement a Transformer-based model for Video captioning, utilizing 3D CNN architectures like C3D and Two-stream I3D for video extraction. We also apply certain dimensionality reduction techniques so as to keep the overall size of the model within limits. We finally present our results on the MSVD and ActivityNet datasets for Single and Dense video captioning tasks respectively.
\end{abstract}

\section{Introduction}
Videos have become synonymous with Information exchange. Every minute, 400 hours of video is uploaded to YouTube, and 46,000 years of video is watched annually. When the number of videos become so huge in size, it becomes a necessity to automatically process these videos. One such way to process these videos is to automatically understand what is the content within them. This would help in automatically tagging them without the need for human effort.

\begin{figure}[h]
\includegraphics[width=\textwidth]
{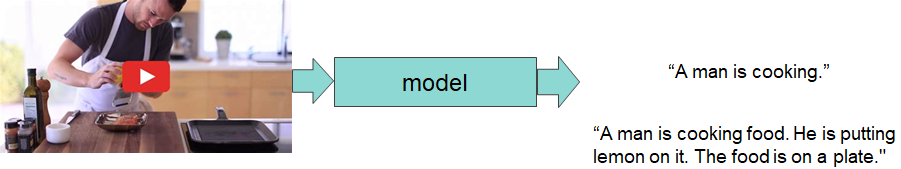}
\caption{Video Captioning and Dense Video Captioning Example}
\label{fig:video-summary}
1

\end{figure}

Video Captioning/Summarization is the process of describing a video in one or more sentences. When more than one sentence is used, it is termed as Dense Video Captioning. A sample is shown in Figure \ref{fig:video-summary}.

\begin{figure}[h]
\includegraphics[width=\textwidth]
{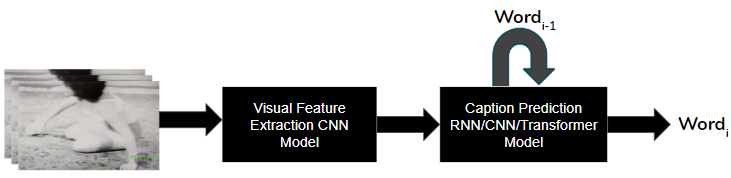}
\caption{Generic Video Captioning Pipeline}
\label{fig:caption-pipeline}
\end{figure}

A generic video captioning pipeline consists of a Video feature extraction network. This network reduces the video dimensions from hundreds of thousands of pixels to only a few thousand floating-point numbers. More on this is covered in Section \ref{video_features} and our implementation of the same is covered in Section \ref{features}. Then, a prediction model is used to generate the captions. Historically, it consisted of Recurrent Neural Network architectures like Vanilla RNNs, GRUs and LSTMs or their variants \citep{Venugopalan2015a, Venugopalan2015b, Pan2017, Donahue2017}. With the recent advances in Attention based mechanisms, soft, hard and self-attention had become the state-of-the-art methods for Video Captioning \citep{Yao2015, Yu2016, Long2018a, Long2018b, Zhou2018a}. A generic video captioning pipeline is shown in Figure \ref{fig:caption-pipeline}.

This report is presented as follows: Section \ref{lit_review} goes in depth into the various Video captioning techniques that have given promising results, Section \ref{architecture} describes in detail our proposed model, while Section \ref{implementation} describes the implementation details of our model, the training and testing procedure used. We then present our results in Section \ref{results}, and Section \ref{limitations} dives into the limitations we noticed with the model and where this project could be heading in the near future.

\section{Literature Review}
\label{lit_review}
Our Literature review is structured as follows - Section \ref{video_features} dives into various networks for video features extraction, Section \ref{transformers} compares vanilla transformers and other augmented networks used for various Sequence Modelling tasks, and Section \ref{video_captioning} explains in detail the different ideas and architectures that have been used for Video Captioning.

\subsection{Video Feature Extraction}
\label{video_features}
The first step in many video analysis tasks including video captioning is extracting features from the raw video input. Because a video can be seen as an ordered collection of images, the majority of video feature extraction methods have been derived from image feature extraction methods. These methods can be divided into three categories, (1) low-level and/or hand-crafted, (2) 2d-CNN, and (3) 3d-CNN. 

Prior to 2013, researches in video feature extraction have focused on adapting handcrafted image feature extraction methods to videos. Among these works, \citep{laptev2005space} extended 2d Harris detector to 3d with time as the third dimension to detect interest points that is conspicuous both independently in each frame and sequentially in time. This method achieved good performance at that time in human action analysis. Another line of methods \citep{barron1994performance} adapts optical flow which is pixel-level gradient aggregated over a local window as a feature extractor for higher level video analysis. In the object tracking community, basic feature extractors are combined with movement modeling to achieve good performance. One prominent example is tracking an object that is known to appear as a "blob" in some form of imaging, either RGB or heatmap, by detecting the largest group of connected bright pixels. Other successful image feature extractors have been adapted to be applied on videos, such as 3d-SIFT \citep{scovanner20073}, and 3d-HOG \citep{klaser2008spatio}. More recently,  \citep{wang2013dense} showed that densely sampling 2d-features combined with optical flow performs well in action recognition. Improved Dense Trajectories(iDT) \citep{wang2013action} which improves upon method proposed in \citep{wang2013dense} by eliminating camera motion in the calculated optical flows thus removing the mismatch between camera motion and object movement. This is the state-of-the-art method in this category of feature extractors. The major drawback of low-level hand-crafted features is their lack of high-level semantic information that is crucial in many video analysis tasks especially video summarization. Moreover, most of such feature extractors are not made to be easily differentiable making it difficult to incorporate them into an end-to-end learning system without sacrificing trainability.

The success of deep convolutional neural networks (CNN) in various computer vision tasks in 2d settings especially object recognition \citep{krizhevsky2012imagenet} spurred the use of CNNs as frame-wise 2d feature extractors in video analysis. A typical CNN consists of several convolutional layers each producing a feature map which captures the hierarchical features of the input image. Variants of CNNs have been proposed to improve performance \citep{simonyan2014very} \citep{he2016deep} \citep{huang2017densely}. A key aspect of CNNs is that the feature maps produced by a task-specific CNN often works well on other tasks, a phenomenon known as transfer learning studied by the machine learning community. Thus, a pre-trained CNN on a related image analysis task can be used as a feature extractor for video analysis tasks. For example, a pre-trained CNN encodes each frame of the video into a feature vector and a sequence modeler such as Recurrent Neural Network (RNN) takes this sequence of features as input. A key characteristic of such approaches is that they extract features from frames independently as  CNN is a 2d feature extractor. They do not take into account the relation between several connecting frames \citep{tran2015learning}. 

Some methods did attempt to utilize CNN's ability of processing 2d images in the form of multiple channels by expanding the input channels from RGB of one frame to multiple RGBs of connecting frames. But the problem with such framework is that the sequential information is lost after one layer, as the convolutional operation squashes input channels from previous layer into one new channel of the current layer. \citep{tran2015learning} proposes to overcome this issue by adapting 3D-CNN to the problem. 3D-CNN expands the same convolutional operation in 2D-CNN to three dimensional space. A sequence of video frames can be seen as a three-dimensional input, 2d in each frame and time as the third dimension. The advantage of 3D-CNN over 2D-CNN is that the convolutional operation produces 3d feature maps which in the context of this problem means that the time dependency between frames is not only retained and modeled over convolutional layers in 3D-CNN. \citep{tran2015learning} shows that their proposed 3D-CNN based video feature extractor C3D outperforms other video feature extractors including 2D-CNN-based feature extractors in action recognition and video object recognition. 

\subsection{Sequence to Sequence models using self-attention}
\label{transformers}
Recurrent Neural Networks (RNN's) and Long Short Term Memory cells have been posed for sequence to sequence tasks for a long time. While RNN's and LSTM's are naturally suited for these tasks, they fail to capture long-term dependencies or adapt to sequence lengths not encountered in training. Machine Translation systems use an encoder-decoder architecture, in which, the outputs of the decoder at each time step are conditioned on the encoder. The inability of RNN's or LSTM's to capture long-term dependencies is well exposed in these encoder-decoder based translation system. The encoded context vector generated by the encoder fails to capture information about the tokens seen in the beginning of the sequence, especially when the sequences are long. This then makes vanilla RNN's and LSTM's unsuitable for modeling translation based tasks having long sequences.

Bengio et al. \citep{BengioNMT} introduced the scaled-dot-product attention mechanism which saw an improvement over the encoder-decoder based architectures. The key idea behind their success was rather than using just the context vector generated by the encoder, soft-attention can help improve the performance of the decoder, by providing it the hidden states of the encoder. In a way, the decoder peeks at the input sequence using an attention distribution to decode the sequence. Several other improvements over this architecture have been proposed, but all of these rely on using either RNN's and LSTM's, and hence, are unable to capture long-term dependencies and cannot be parallelized across training examples.

The transformer model \citep{Vaswani2017} addressed the shortcomings of recurrent machine translation systems by proposing an architecture that relies only on self-attention. Since they take recurrence completely out of the picture, the Transformer model allows parallelization across training samples and generates a feature representation in a fixed number of steps, which are chosen empirically. They also use the scaled dot product attention mechanism over the Keys K, Queries Q and Values V and compute the representation using softmax as:

$$
    Attention (Q, K, V) = softmax(\frac{QK^T}{\sqrt{d_k}}) V
$$
Vaswani et al. \citep{Vaswani2017} use N = 8 attention heads and concatenate the outputs of each of these heads to compute the self-attention representation of the inputs. These representations are then fed to a point-wise feed forward neural network to generate the final encoded context vectors. Their complete architecture thus uses self-attention for the inputs, encoder-decoder attention for the decoder and masked self-attention to generate the outputs.

$$    MultiHead (Q, K, V) = concat(head_1, head_2, ....., head_N) W_O
$$

While the Transformer improves upon the vanilla RNN and LSTM based models, it fails to generalize to unseen input lengths or learn simple tasks like Copying and Repeat copying. The Universal Transformer\citep{Universal} model aims to address these issues by weight sharing across the encoder and decoder units and by using Adaptive Computation Time \citep{ACT} to learn the number of steps required to learn the encoder representation. Adaptive Computation Time is an approach using which sequence models can dynamically learn the number of computation steps required to process an input. Earlier, these steps had to be explicitly defined by the architecture or were dependent on the input lengths. The Universal Transformer model uses ACT not between inputs, but across depth, and refine their self-attention distributions dynamically. 

The Universal Transformer uses 4 attention heads instead of the 8 as proposed in the original architecture, and achieve significant improvements over the vanilla transformer. Their adaptation of the ACT method can indeed prove to be significant when dealing with videos of large and varying lengths. The number of parameters for both these models is the same, which has prompted us to believe that this architecture can indeed do better for tasks like video summarization, under the same computational constraints.

There have been a few other notable improvements over the vanilla transformer architecture. Transformer-XL networks \citep{TransformerXL} bring back recurrence in the transformer models and counter the problem of fixed-length context in the Transformer. They use recurrence at a segment-level and show that the Transformer-XL model can capture long-term dependencies. They also demonstrate that by doing this, they achieve a 1800\% improvement in the evaluation time when compared to the Transformer model. BERT \citep{BERT} uses bidirectional Transformers and condition the representations on both left and right context, for all layers. They propose a pre-training scheme which yields state-of-the-art results across a wide variety of language modeling tasks.

\subsection{Video Captioning}
\label{video_captioning}
Video Captioning has always been seen as a Sequence Modelling problem. Before Recurrent Neural Networks came into prominence, Hidden Markov Models were popular for tackling such tasks. \citep{Yu2013} used an object detector to detect objects within the scene and track them from one frame to another. They also detected pose, shape and view-point specific features within the video and extracted colour, shape and size specific features for each frame. With the help of these features, they try to model lexicons like nouns, verbs, adjectives and adverbs using Hidden Markov Model. For example, verbs like \textit{jump} and \textit{pick-up} are represented as a two state HMM over velocity features; nouns like \textit{person} is represented as a one state HMM over image features; and adjectives like \textit{red} and \textit{big} are modelled with a one state HMM over the relative position of the objects. The authors have also given examples of sentences modelled together by the above defined HMMs. Once the features are available, HMMs determine the lexicons in the available words and frame the sentence. The authors conducted real-world, albeit limited experiments consisting of a person, backpack, a chair and a trash-can, and the models output the sentence from the lexicons pairings it had learned.

By 2015, with the resurgence of LSTMs, and introduction of GRUs and Attention mechanism, RNNs had become synonymous with video captioning. Yao et. al \citep{Yao2015} used a 3D CNN-RNN encoder-decoder architecture to capture spatio-temporal information. Work prior to this had used an object detector like VGG-16 to extract features and a 2 layer LSTM network for caption prediction \citep{Venugopalan2015a}. \citep{Yao2015} proposed to use a 3D convolutional network on a cuboid of feature block obtained by computing the Histograms of Oriented Gradients, Optical Flow and Motion Boundary. This pre-processing was done so as to reduce the execution time of the 3D CNNs, which consisted of three 3D convolution layers followed by ReLU activation, pooling and fully connected layers. A soft attention is applied to the output of the fully connected layer which is then passed to a single layer LSTM network for caption prediction. The authors tested their model on the Youtube2Text dataset and achieved state of the art BLEU, METEOR, CIDEr and Perplexity scores on the dataset. The authors, however, did not test on the more complex COCO dataset which they have proposed as their future work.

While the above paper \citep{Yao2015} introduced the concept of using 3D convolutions to extract spatio-temporal features from video, it still relied on hand-crafted features like HOGs, HOFs and MBH. Another similar, but equally influential paper in this field is \citep{Venugopalan2015b}. They proposed multiple models based on the encoder-decoder architecture consisting of a 2-layer LSTM network. The input to the encoder was computed from, (1) a pre-trained VGG network with RGB frames as input, (2) AlexNet with optical flow images as the input which is pre-trained on the UCF-101 dataset, and (3) VGG model with both RGB frames and optical flow images as the input. The output of the decoder predicts the caption of the incoming video. The optical flow images were computed using a classical technique described in \citep{Brox2004}. The model which uses both RGB and Optical flow images gave state of the art METEOR score on the MSVD dataset, thus concluding that spatio-temporal features are better at capturing video features. They also evaluated their model on the more challenging MPII-MD and M-VAD movie description datasets and achieved promising results. 

Both the previous papers were not end-to-end trainable due to the pre-processing done in extracting HOG, HOF, MBH and optical flow features. \citep{Donahue2017} introduced a recurrent convolutional network, an end-to-end deep network based on feature extracting convolutional network to account for the spatial domain, and a recurrent network based on LSTMs to model the time domain. The authors claimed that their model posses long-term memory and can be adapted to a variety of sequence modelling tasks. They tested their model on Activity Recognition, Image and Video Captioning datasets. For Video captioning specifically, they used CRF to model various components of the input video. These CRF features are then one-hot encoded and fed into, (1) LSTM based encoder-decoder network or, (2) LSTM based decoder only network, for video description. The authors have shown their model to achieve a better BLEU score than other comparable models on the TACoS multilevel dataset. Their main contribution is designing a network which gives promising results for three diverse sequence modelling tasks, however, they have noted that incorporating temporal features from the video and using an attention model could significantly improve the performance. 

Donnahue et al. \citep{Donahue2017} have shown how merged deep CNN and RNN architectures can be utilized for this task. \citep{Yu2016} followed a similar approach in their a Heirarchial RNN network, which consisted of the following parts - (1) A video descriptor network for modelling the spatio-temporal features, (2) An attention mechanism for selecting the most suitable video features, and (3) A multimodal layer to incorporate both video and text features. For the video descriptor layer, they proposed three methods. The first method included using a pre-trained VGG network to extract features, the second method involved using a pre-trained C3D network for spatio-temporal features and the last method was to compute optical flow images from the input video and then use a pre-trained VGG to extract the features. These video features were then fed into a soft attention layer after which they are combined with the hidden layer of the GRU network to which an embedding of the text served as the input. After the combination in the multimodal layer, which multiplies each of the input with separate weight matrices, a hidden and a softmax layer is used to generate the words. The authors have also modelled another GRU layer, which they call paragraph generator. The role of this layer is to receive the context vector and the word embeddings, and maintain a semantic context, which will be used to initialize the first recurrent network on arrival of the next input. Their optical flow based h-RNN has given state of the art performance on the YouTubeClips and TACoS multilevel datasets on BLUE, CIDEr and METEOR scores, closely followed by the C3D based model. The authors have also pointed out that they used C3D model trained on the Sports-1M dataset which has videos that are quite different from the ones they trained and tested the model on. Also, they have noted that their method failed to incorporate objects that were very small in size. Another drawback of this model is that since the paragraph generator which captured the context is used to initialize the main recurrent sentence generator, any error in this would be propagated through the network.

Inspired by the above approaches of (1) using CNN-RNN hybrid architecture and (2) using text input as well, \citep{Pan2017} have proposed an LSTM based Transferred Semantic Attributes model (LSTM-TSA) which learns semantic attributes from the videos and image frames and inputs them to the LSTM layer for caption generation. They adopt the image semantic attributes detection framework of \citep{You2016} to videos and generate words that lie in the ground truth caption of the video. Such image and video attributes respectively are computed from pre-trained VGG network on the ImageNet dataset and C3D network pre-trained on the Sports-1M dataset. Then these individual attributes are passed through gated function, the output of which is multiplied with these individual attributes, and then passed to an LSTM layer. Their LSTM-TSA model has given the best BLEU, CIDEr and METEOR scores on the MSVD dataset. They have also compared their model with others on the M-VAD and MPII-MD datasets, on which they have shown very promising results. As a future work, the authors have noted that they want to incorporate attention mechanism into their framework to focus on essential parts, which they believe can achieve better results.

Some of the previous work described here have either demonstrated or noted that attention mechanism can significantly improve Video captioning results. Therefore, most of the research happening currently use some form of attention mechanism. As a result, most of the research has been focused extensively on incorporating attention with video captioning. \citep{Long2018a} have used multi-modal keyless attention for video classification with (1) input RBG frames features, (2) optical flow images, and (3) acoustic features by formulating Mel-spectrogram images. The features of all the three types of inputs are extracted using VGG which was pre-trained on the ImageNet dataset. All these features are 1D max pooled to obtain a lower dimensional set of attributes which are then passed onto a 2-layer Bi-directional LSTM network. The output of this network is then fed into an attention mechanism, which they call keyless attention, similar to a 2-layer feed forward network with a softmax layer at the end, to compute the attention weights. The output of their Keyless Attention is fed into a feedforward network to compute the video class probabilities. They have tested their model on the YouTube-8M dataset and predicted multiple tags for each of the videos in the dataset. Although this paper did not directly use the model for video captioning, this paper was included due to their novelty in incorporating sound along with the video information and usage of attention mechanism. 

Same authors of the previous paper have also proposed another model based on multi-modal attention mechanism for video captioning \citep{Long2018b}. In this paper, instead of audio features, the features used consisted of embeddings from text, frame-level features extracted from a pre-trained ResNet-152 model and video level motion features from a pre-trained C3D model. All these features are passed into separate attention layers so as to focus on important features individually; then the concatenated output is made to pass through a single-layer LSTM network whose hidden state is then attended to and passed to a softmax layer to predict the captions. The authors compared their model on MSVD and MSR-VTT datasets. These datasets do not contain any semantic information. To deal with this, the authors trained a ResNet-152 network on the COCO dataset predicting multiple labels for the captions which is then fed as input text to their attention network. Therefore, the previous three papers have shown that incorporating features other than video, like audio or text can significantly improve the model's performance.

All the above papers which relied on attention, used a soft attention model or a feedforward model. \citep{Zhou2018a} have used an augmented vanilla transformer with self attention, which consists of one encoder and two decoders. The input to the encoder is the features extracted from RGB images and optical flow images computed using a ResNet-200 network which is pre-trained on ActivityNet dataset. This forms the input to the encoder of the multi-headed self-attention layer in the transformer. The first decoder, which the authors term as proposal decoder is based on ProcNets \citep{Zhou2018b} and outputs regions of a video comprising of similar context which can be explained by a single sentence. It uses anchor boxes concept of Object Detection applied in the temporal direction, to output such segments of the video which are deemed as important. Next, this input, along with the self-attended output of the encoder is passed to a second decoder, named  captioning decoder. Using self-attention, it captures the important information from the encoder output for each of the proposed segment by maintaining a masking function. The output of this decoder are the captions to be predicted. The authors tested their model on the ActivityNet Captioning and YouCookII datasets obtaining encouraging results, which were significantly better than the RNN counterparts, thus demonstrating the advantages of using transformer over recurrent networks.

\section{Architecture} 
\label{architecture}

\subsection{Video Feature Extraction}
\label{features}
\begin{figure}[h]
    \centering
    \includegraphics[width=\textwidth]{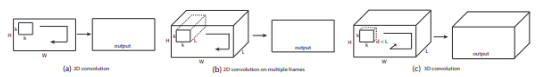}
    \caption{3D convolutions preserving spatial as well as temporal information present in a video}
    \label{fig:3d-conv}
\end{figure}

Instead of using frame-level feature extractors, we use networks which give us spatio-temporal features directly from videos. These architectures use 3D convolutions to encode spatial as well as temporal information present in videos. As highlighted in Figure \ref{fig:3d-conv}, using 2D convolutions on an image or a video (set of frames) result in a single feature map. However, using 3D convolutions on a set of frames result in a set of feature maps. The number of feature maps depend on the size of the temporal kernel and the strides used.

\begin{figure}[h]
    \centering
    \includegraphics[width=\textwidth]{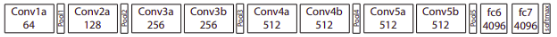}
    \caption{C3D architecture as introduced by  Tran et al. \citep{tran2015learning}}
    \label{fig:c3d}
\end{figure}

\begin{figure}[h]
    \centering
    \includegraphics[width=\textwidth]{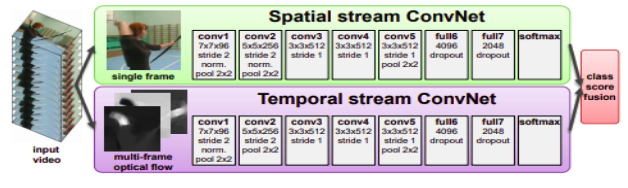}
    \caption{I3D architecture uses two ConvNets running in parallel over the RGB stream and the Optical Flow stream}
    \label{fig:i3d}
\end{figure}

Recent advancements in the field of activity recognition have brought about various architectures which can serve as good spatio-temporal feature extractors. We look at architectures that can provide temporal information directly, instead of relying on a recurrent network to encode information from each time step. For this specific feature extraction task, we use C3D (3D Convolutional Neural Networks, Figure \ref{fig:c3d}) and I3D (Inflated 3D Convolutional Neural Networks for Activity Recognition, Figure \ref{fig:i3d}) to extract features for the Transformer model. I3D was inspired by the popular two-stream architecture for video classification, which has two similar networks running on the RGB stream and the Optical Flow stream. For I3D, we take the output from $Mixed_{5c}$ layer, which gives a feature vector of length 1024 for each 8 frames. For the C3D architecture, we take the output from $fc_6$ layer which gives a feature vector of length 4096 for each 16 frames.

\subsection{Transformers}
The transformer architecture is a rich and expressive model capable of producing state-of-the-art results on a wide variety of language modeling tasks. However, to the best of our knowledge, this is the first work which explores the capability of transformers to learn captions including full paragraphs. In order to apply self-attention for videos, we had to make a few notable changes from the original architecture. Since we already have feature representations for each time step, we skip the embedding layer used in transformers. Note that the original transformer model learns this embedding layer during training, which as one would guess, results in significant improvements compared to using a frozen representation. For a task like video summarization, this would mean learning or fine-tuning the feature extraction layers as well. However, due to limited compute resources, this was not done for now and remains as an essential improvement to explore for us in the future. A detailed explanation of the architectural changes from the original model is presented in the next section.

\subsection{Universal Transformers}
Transformers, being such a rich and expressive model, require a lot of data to train. Hence, it is not surprising to see that they fail in a lot of simple algorithmic and memory tasks, as pointed out by Dehghani et al \citep{Universal}. Since we want to experiment with a few datasets which are not that large in size, Universal Transformers are a natural choice. They tie weights of all the encoder and decoder layers present in the transformer model, and use dynamic halting by introducing Adaptive Computation Time \citep{ACT}. A detailed explanation of the architectural changes made for this specific task is provided in the following section.

\begin{figure}[h]
    \centering
    \includegraphics[width=\textwidth]{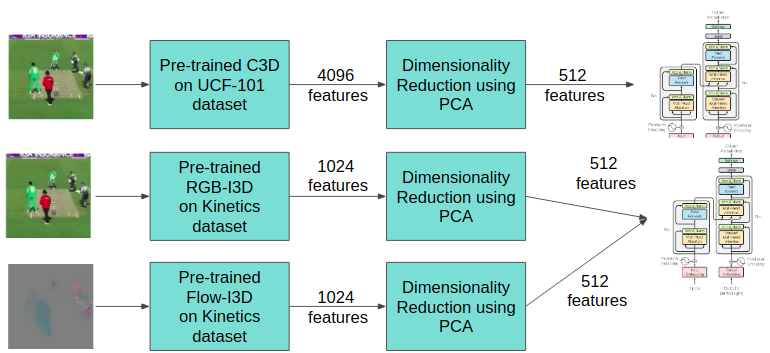}
    \caption{Pipeline for Video Summarization using C3D and I3D as feature extractors and Transformers as a sequence 2 sequence model}
    \label{fig:architecture}
\end{figure}

The detailed architecture incorporating the above networks is shown in Figure \ref{fig:architecture}.

\section{Implementation Details}
\label{implementation}

This section provides the procedure used for training and testing our model. We tested our model on the MSVD dataset, which is used to generate a single caption for each video, and the ActivityNet dataset, which is used to generate dense video captions. Each of the following sub-sections describe the dataset in detail along with the implementation details.  We coded the models in PyTorch by partially adapting a publicly available PyTorch implementation of the original Transformer model (https://github.com/SamLynnEvans/Transformer).

\subsection{MSVD}
\subsubsection{Dataset}
The Microsoft Video Description dataset (MSVD) consists of 1970 video clips of length 10s to 25s obtained from YouTube with subjects being humans and animals. Out of the total, 1300 videos are used for training and the rest 670 are used for evaluation. This is a fairly small dataset with near constant semantics, with most videos consisting of humans performing some activities. The descriptions are produced by humans in multiple languages, with an average of 41 descriptions generated per video. Out of the total descriptions, there are 85,000 descriptions in total for English. These English descriptions together constitute a vocabulary of length 14,000 words. This is a commonly used dataset for video captioning and BLEU is the most common metric used for evaluation.

\subsubsection{Data Preprocessing}
As discussed in Section \ref{features}, we use C3D and I3D as our feature extraction networks. 

In the case of C3D, the $fc_6$ layer of the network pre-trained on the UCF-101 \citep{Soomro2012} was used. For the input, the number of frames to be used was capped at 500, since it was not possible to load more than 500 frames for generating the features on a single GPU. The batch size was taken to be 1, which means that all the frames from a single video served as the input in each iteration of the extraction procedure. C3D generates 4096 feature vector per 16 frames input to the network, hence, in our case, the maximum dimension of the features was 31 x 4096. Finally, Principal Component Analysis, a technique used to reduce the feature size for videos \citep{caba2015}, was used to reduce the feature dimension to 512. This was then stored as a numpy file to be used for caption prediction.

In the case of I3D, the $Mixed_{5c}$ layer was taken as the feature extraction layer, followed by an average pool. The network was pre-trained on the Kinetics dataset \citep{kay2017}. To reduce the memory footprint on the GPU, the number of frames were capped to 400. This is smaller than 500 frames used for C3D as I3D is a bigger and deeper network. The batch size was again taken to be 1, and the output generated were 1024 feature vector for every 8 frames. Again PCA was applied to reduce the dimension of the features to 512. Hence, the maximum feature size stored as a numpy file was 50 x 512. I3D, as stated before, can be used as a two-stream network as well. Therefore, we also extracted Optical FLow features using Farneback's dense optical flow features \citep{Farnebck2003}. Although the original Kinetics dataset used TV-L1 method for flow estimation \citep{Prez2013}, we found that without CUDA support in OpenCV, it was taking 15s to 20s to compute for a single pair of image, hence, we went with the former method. Similar to RGB feature extraction, 400 flow frames of channel depth 2, were used for extraction. This resulted in a maximum feature vector of 50 x 512 size after application PCA.  

In both the above cases, the input images were normalized and center-cropped before feeding into the network.

\subsection{ActivityNet}
\subsubsection{Dataset}
The ActivityNet dataset \citep{caba2015} consists of 20,000 videos obtained from YouTube with people performing certain activities. These activities involve dancing, cooking, speaking, among others and on an average are 180s in length. ActivityNet is widely used for (1) Activity Classification, (2) Event proposal detection, and (3) Dense Video Captioning. We have used it for the latter, wherein a paragraph comprising of 3 to 4 sentences is generated per video. The dataset on an average consists of 100,000 video descriptions with an average of 13.68 words per sentence and 3.65 such sentences per video, with the total length of vocabulary being 13,300 words. The authors also split the dataset into training, validation and testing sets, however, for dense video captioning, the validation set is used for evaluating the network as done in \citep{Zhou2018a}. This is because the testing script provided by the authors expects the localization as well, and the testing ground truths are not made available in the dataset. Finally, BLEU score is the most popular metric for caption prediction quality estimation.

\subsubsection{Data Preprocessing}
The ActivityNet dataset provides C3D features which they compute on the videos of the dataset. These C3D features have a feature dimension of 500, and the features are extracted from the $fc_6$ layer to which PCA is applied. Since, there is no video available in the dataset, only C3D features were used for dense video captioning.

\subsection{Caption Prediction}
For both the above datasets and feature inputs, similar configurations for the Transformer and the Universal Transformer were used. For the MSVD dataset, since the number of descriptions per video was high, the pairing of the video was done randomly with one of the available captions. This was done to increase the number of available pairs of videos and captions. For AcivityNet, only one paragraph was available for training. In all the combination of features and models, the batch size used was 64.

In both the cases of Transformers and Universal Transformers, the embedding layer of the Encoder was removed since the inputs to the networks were no longer the semantic words. Other changes made included the altering of the learning rate schedule. Both of the transformer models used a learning rate schedule called 'CosineWithRestarts'. This scheduler basically increased the learning rate linearly during the warm-up stage, and then reduced proportionately with increasing number of epochs. The default learning rate proved to be too large for our type of input and hence, as a result, the model diverged after a certain number of epochs. Therefore, we used a uniformly reducing learning rate with the decreasing factor of 0.98, which helped in better learning without causing any divergence. 

For the MSVD dataset, the number of layers used for the transformer was 6, with the dimension of the model being 512 and the number of multi-attention heads being 8. In case of Universal transformers, the number of layers used were 8 of 512 dimensional each, with 8 multi-attention heads forming the self-attention layer. Since we used features extracted by I3D and C3D networks, and not the learned embedding representations during training, this made the task a bit more challenging. We noticed that by incorporating adaptive computation time, the universal transformer was halting way too early for both short and longer videos. This greatly reduced the capacity of the model and we decided not to incorporate ACT in Universal Transformers. Even without ACT, the universal transformers took considerably shorter time and memory to train. 

Seeing the qualitative results by inspecting the captions, we felt that Universal Transformers gave more diverse results. Hence, we decided to train the ActivityNet only using the Universal Transformer model. Due to the size and complexity of the ActivityNet dataset, a number of changes were made to the model. The models' dimension was changed to 500 to match with the number of input features of ActivityNet C3D extractor, and the multi-attention heads were also changed to 10 so as to make it perfectly divisible with the dimension of the model. Also, 8 Encoder-Decoder layers were used instead of the 4 used in the original paper.

\subsection{Image Attributes Generator} 
\label{annotation}
As detailed in Section \ref{results} we faced the problem with the nouns getting mixed up in the MSVD dataset, and hence, taking cues from previous research \citep{Pan2017,Long2018a,Long2018b}, we implemented an image annotation network which predicted the annotations for an image. These annotations would serve as the text features to the transformer network. For this, we trained a ResNet-50 model and a VGG-19 model, both pre-trained on the COCO and the ImageNet dataset respectively. For the training, we selected the 10 most frequently occurring words among the descriptions available, which would serve as the ground truth words for the input video frames. A fully connected layer was added at the end whose input was the attended weights across all frames. Let $v_i$ be the features from the individual frames, then the input to the last fully connected layer ($fc_{last}$) was - 
\[\alpha_i = \frac{\exp{v_i}}{\sum_{i=1}^{n} \exp{v_i}}\]
\[v_{in} = \sum_{i=1}^{n} \alpha_i v_i\]
\[y = fc_{last}(v_{in})\]
However, our findings were in contrast to the papers cited above. Both the ResNet-50 and VGG-19 failed to generate diverse set of words for the validation set. The words generated were those that were already predicted well by the transformer model. So, adding this generator did not prove to be of any help. It had also failed to capture the nuances in the dataset.

\section{Results} 
\label{results}

We present both quantitative and qualitative results of our model on the two datasets. It is standard to use BLEU score as a quantitative measure for video summarization tasks \citep{papineni2002bleu}. BLEU is originally developed for machine translation but has adapted over the years to similar text generation tasks such as image captioning and video summarization. Evaluating machine generated text quantitatively is still a problem to this day, but since BLEU is a widely used metric reported by state-of-the-art methods that we compare against, we calculated BLEU score on the results of our model as a quantitative evaluation. We also take a qualitative look at the results and showcase some examples where our model performed poorly to uncover the limitations of our model.  

\subsection{MSVD}
Our model is able to achieve very promising BLEU scores on MSVD dataset. Table \ref{table:Table 1} shows that our model performs at the same level as the state-of-the-art models at the widely used BLEU-4 metric while outperforming the state-of-the-art models at BLEU-1 and BLEU-2. BLEU-1 and BLEU-2 mostly represent correctness at word and 2-word phrase level while BLEU-4 characterizes average correctness at word level and up to 4-word phrases. Our model has higher BLEU-1 and BLEU-2 indicating that it is better at producing the correct words for captions, especially nouns and verbs which are key to video summarization. This could be the effect of Transformer model in general where it can selectively pay attention to any frame of video feature while generating any word regardless of position. Furthermore, our model's BLEU-4 score that is at the same level as state-of-the-art shows that not only can our model generates the correct words, it can also produce correct meaningful phrases and sentences. 

The results are obtained by giving the model visual inputs only. The Attributes Generator model did not help us in improving the performance, as discussed in Section \ref{annotation}. We still feel that it is important to compare our model with other state of the art methods which used both visual inputs and semantic inputs. Both the LSTM-TSA proposed in \citep{Pan2017} and Multi-faceted Attention model proposed in \citep{Long2018a} have noted that they are able to significant increase performance of their model by combining semantic features. However, same was not observed in our case.

\begin{table}[ht]
\caption{MSVD BLEU scores. Our model outperforms the state-of-the-art models on MSVD dataset in BLEU-1 and BLEU-2. Our model performs a the same level as the state-of-the-art models in BLEU-3 and BLEU-4.}
\centering 
\begin{tabular}{l|l|l|l|l}
 \hline
 Model & BLEU-1 & BLEU-2 & BLEU-3 & BLEU-4 \\ \hline  \hline
 LSTM-TSA \citep{Pan2017} & 0.828 & 0.720 & 0.628 & \bf{0.528} \\
 Multi-faceted Attention \citep{Long2018b} & 0.830 &  0.719 & \bf{0.630} & 0.520 \\  \hline
 C3D + Transformer (Ours) & 0.906 & 0.762 & 0.621 & 0.517 \\
 I3D + Transformer (Ours) & 0.889 & 0.731 & 0.564 & 0.442 \\
 C3D + Universal Transformer (Ours) & 0.901 & 0.765 & 0.587 & 0.501 \\
 I3D + Universal Transformer (Ours) & \bf{0.910} & \bf{0.782} & 0.521 & 0.460 \\ \hline 
\end{tabular} 
\label{table:Table 1}
\end{table}

\begin{figure}[h]
\includegraphics[width=\textwidth]
{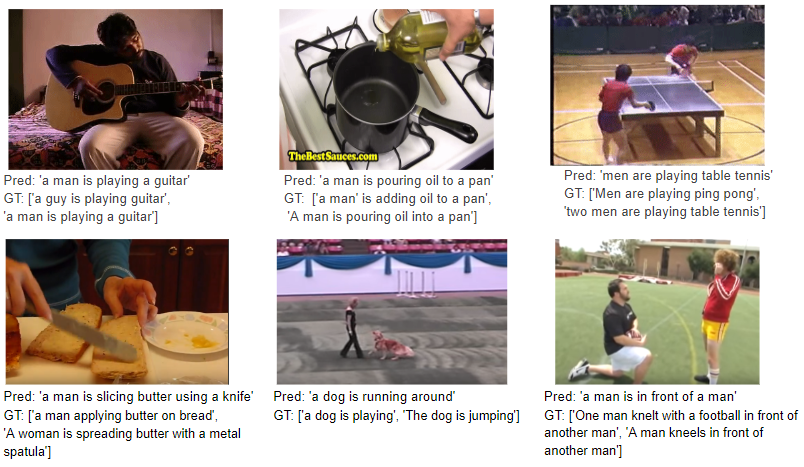}
\caption{Positive Examples from MSVD}
\label{fig:msvd-positive}
\end{figure}

We also conducted qualitative examination of the results. Figure \ref{fig:msvd-positive} shows some examples from the test set where our model (C3D + Universal Transformer) performs reasonably well. Both the subject present in the video and the activity of that subject are correctly identified and output in a semantically correct sentence. However, we also observed some examples where our model did not perform so well as shown in Figure 8. Either the subject or the activity is not identified completely correct. However, the model is still able to extract some meaningful information from the videos. For example, in the walking turtle video, the subject, i.e. turtle, is not correctly identified, but its action, i.e. walking, is correctly identified. 

\begin{figure}[h]
\includegraphics[width=\textwidth]
{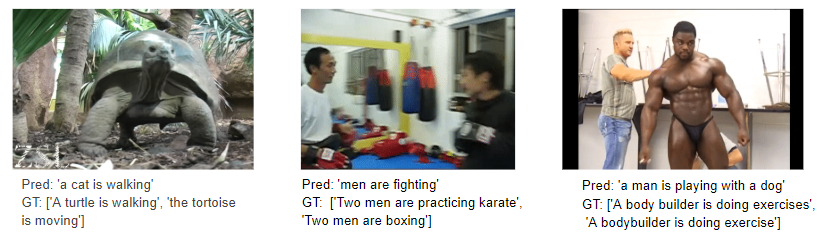}
\caption{Negative Examples from MSVD}
\label{fig:msvd-negative}
\end{figure}

We observed that MSVD dataset lacks diversity. As mentioned earlier, the pre-processed vocabulary from description texts reaches 14,000 while the dataset only contains less than 1,970 videos. This means that many objects and activities only appear in one video, which often forces the model to overfit as we have observed. For example, the turtle video in Figure \ref{fig:msvd-negative} is the only video in the dataset that contains a turtle. Some of our models did also output the same sentence many times to similar looking videos, which shows that the model only learned the text semantics or just memorized descriptions and not the task we intended to solve, which is to somewhat understand and describe videos with texts. Such results would still obtain a moderate BLEU score due to the text semantics present in the generated texts. It raises the question of whether or not the results obtained on the MSVD dataset can be generalized. To further test our model, we used a much more complex dataset, namely, ActivityNet.

\subsection{ActivityNet}
Our model gives promising paragraph-wise BLEU score on the ActivityNet dataset as shown in Table \ref{table:Table 2}. It should be noted that our results for the ActivityNet Dense captioning cannot be directly compared with \citep{Zhou2018a}. The purpose of our task was to demonstrate that our model can also be used to generate a multi-sentence summary of the video, rather than just a single line caption, without any modifications to the network. By this we felt that we would be really pushing the boundaries of our model by allowing it to learn on its own the important aspects of the video. Hence, we do not make use of any other information other than the video features provided in the dataset. In comparison, \citep{Zhou2018a} have used the ground-truth event proposals to generate a single caption from a particular proposal, and they are able to generate a paragraph based on 3 to 4 ground-truth event proposals given for each video. This distinction is important because generating multiple sentences sequentially to describe a long video without any additional input is a relatively difficult task in the video summarization realm. This could be an important topic for future research. To our best knowledge, our model provides the first BLEU score benchmark in this task.

\begin{table}[ht]
\centering 
\caption{ActivityNet BLEU scores. Our model gives state-of-the-art paragraph-wise BLEU score on the ActivityNet dataset.}
\begin{tabular}{l|l|l|l|l}
 \hline
 Model & BLEU-1 & BLEU-2 & BLEU-3 & BLEU-4 \\ \hline  \hline
 I3D + Universal Transformer (Ours) & 0.710 & 0.659 & 0.577 & 0.490 \\ \hline
\end{tabular} 
\label{table:Table 2}
\end{table}

\begin{figure}[h]
\includegraphics[width=\textwidth]
{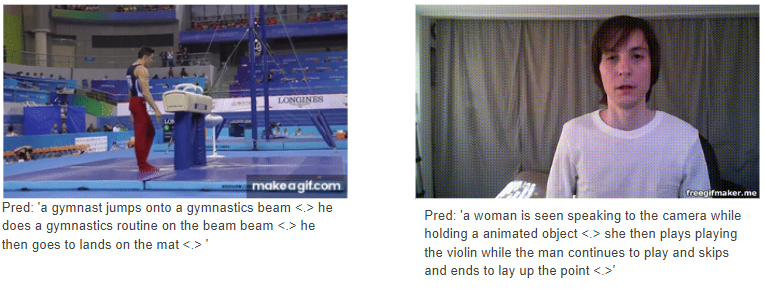}
\caption{Positive Examples from ActivityNet}
\label{fig:activitynet-positive}
\end{figure}

\begin{figure}[h]
\includegraphics[width=\textwidth]
{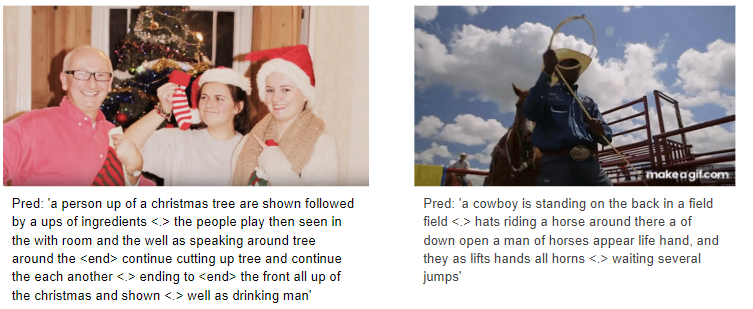}
\caption{Negative Examples from ActivityNet}
\label{fig:activitynet-negative}
\end{figure}

Some examples where our model (I3D + Universal Transformer) performs reasonably well are shown in Figure \ref{fig:activitynet-positive}. The model is able to identify the subject correctly and the sequence of actions the subject took. It is important to note that in the gymnast example, at lest 50 percent of the video does not contain any meaningful information. Many examples are similar where a substantial portion of frames do not contain activity information to be described, so our model learned to both pay attention to frames that need to be described along with how to describe. This evidently shows the advantage of Transformer models in video summarization when a relatively large portion of the video is irrelevant. Some examples where our model performs poorly are shown in Figure \ref{fig:activitynet-negative}. The model can correctly identify the subject and the activity, but could not put the activity sequence into meaningful sentences and form a coherent paragraph. 

From qualitative examination of both datasets, we conclude that Universal Transformers gives a diverse set of captions than Transformers, despite similar BLEU scores, probably due to its simpler structure due to weight sharing that prevents overfitting to ground-truth text descriptions.

\section{Limitations and Future Work}
\label{limitations}
Some notable limitations of our model include nouns getting mixed up, activities not correctly identified, and the the failure to form coherent paragraphs after identifying the nouns and activities. However, it's unclear to what degree can these drawbacks be improved by hyper-parameter tuning and fine-tuning on pre-trained networks. Another major drawback of our models are their inability to incorporate image attributes which might require proper hyper-parameter tuning and/or exploration of other networks for this task. Thus, as a short-term plan (2 weeks),  we would want to focus our immediate attention on hyper-parameter tuning and fine-tuning the networks which include, C3D and I3D feature extraction networks, the Image Attribute Generator, as well as the Transformer and Universal Transformer models themselves. Fine-tuning C3D and I3D feature extraction networks with the task training set instead of using the trained networks as they are provided has shown to improve performance on various other video tasks in the past. This would require slightly more powerful or another GPU due to the size of these 3D CNN networks. We also plan to further refine the model so as to be able to capture the nuances, particularly in the dataset like the MSVD which has constant temporal semantics. Hence, as a long-term goal (4 weeks+), we would like to try ACT in Universal transformer to improve the halting process, which have been halting very early during the training process. Another thing we would want to try is to explore other Image Attributes Generator networks, especially the ones that go beyond the conventional CNN networks, as proposed in \citep{Wu2015, Liu2017}. The predicted attributes, if generated correctly, could help in capturing the nuances in any dataset, as it has been explained by other authors.  

\section{Conclusion} 
\label{conclusion}
In this project, after first reviewing the various methods that have been used for Video Captioning, we presented a model based on Transformers for the same. Our implementation makes use of C3D and Two-stream I3D feature extraction networks which serve as the input to the caption prediction network. We then demonstrated the usefulness of our modified Transformer and Universal Transformer models for the video caption prediction task. After showing the success of our model on the MSVD dataset, we further tried to push the limits by testing it on the more complex Dense Video Captioning task and on the much larger ActivityNet dataset. The results of our experiments show that Transformer-based networks, along with the 3D spatio-temporal video feature extraction networks, can achieve great results without the need for any other form of input, and in some cases can even beat the previously attained BLEU scores. We feel that with the ideas presented in Section \ref{limitations}, our model can achieve even more promising results for both single and dense video caption prediction tasks.


\end{document}